\documentclass{article}
\usepackage{amsmath,graphicx,mlspconf}
\usepackage{algorithmic}
\usepackage{graphicx}
\usepackage{textcomp}
\usepackage{xcolor}
\usepackage[caption=false]{subfig}
\usepackage{float}
\usepackage{amsmath,amssymb,amsfonts}
\usepackage{cite}
\newcommand {\vc} [1] {\boldsymbol{#1}}

\title{Self-Compression in Bayesian Neural Networks}
\name{Giuseppina Carannante, Dimah Dera, Ghulam Rasool and Nidhal C. Bouaynaya }
\address{Rowan University,
 Department of Electrical and Computer Engineering, 
Glassboro, NJ\\
carannang1@rowan.edu, derad6@rowan.edu, rasool@rowan.edu, bouaynaya@rowan.edu}

\begin{document}

\maketitle

\begin{abstract}
Machine learning models have achieved human-level performance on various tasks. This success comes at a high cost of computation and storage overhead, which makes machine learning algorithms difficult to deploy on edge devices.  Typically, one has to partially sacrifice accuracy in favor of an increased performance quantified in terms of reduced memory usage and energy consumption. Current methods compress the networks by reducing the precision of the parameters or by eliminating redundant ones. In this paper, we propose a new insight into network compression through the Bayesian framework. We show that Bayesian neural networks automatically discover redundancy in model parameters, thus enabling self-compression, which is linked to the propagation of uncertainty through the layers of the network. Our experimental results show that the network architecture can be successfully compressed by deleting parameters identified by the network itself while retaining the same level of accuracy.

\end{abstract}
\section{Introduction} \label{sec:intro}

Deep Neural Networks (DNNs) have achieved outstanding results in a variety of domains, including computer vision and natural language processing \cite{karras2019style, radford2019language}. One of the key factors responsible for these successes is the ability of modern DNNs to learn a large number of parameters \cite{radford2019language}. However, an increase in the number of parameters is directly linked to higher computational complexity, the requirement for substantial computational resources, and additional storage for the learned parameters. The availability of dedicated computational hardware (e.g., GPUs) and novel training techniques (e.g., distributed training) have stimulated and supported the growth of DNNs. However, resource-constrained systems, including mobile and edge devices (with limited storage and energy), cannot support these DNNs \cite{raina2009large, dean2012large}.

Recently, research communities have focused on exploring compression techniques to reduce the memory usage and computational requirements of DNNs. Some well-known current strategies include \textit{pruning} and \textit{quantization} of neural networks. \textit{Pruning} aims at eliminating (most) redundant connections from the network, while \textit{quantization} is characterized by the use of low-precision representation for parameters and activations. Traditionally, most implementations of DNNs used single-precision floating-point format (FP32). However, recently, half-precision (FP16), and low-precision including INT8, INT4, and binary formats have been explored \cite{courbariaux2015binaryconnect}.  

Various quantization techniques have been proposed to make DNNs perform faster and fit larger networks on edge devices with limited storage capacity and energy budget \cite{goncharenko2018fast, choukroun2019low, rastegari2016xnor}. An unfortunate consequence of quantization is the reduced accuracy, which can be tackled by increasing the network size, performing quantization only on parameters (and not on activations), or fine-tuning and re-training the network. Although many real-world applications can benefit from various quantization techniques, the reduced accuracy offsets the benefits of low precision and significantly increasing training and testing time.  

Pruning techniques reduce the total number of parameters and corresponding computations using an ``importance'' measure. The pruning process is based on the assumption that the trained networks may have redundant parameters that do not contribute to the output of the network and can be removed without compromising the performance \cite{han2015deep}. Parameter pruning techniques are generally aimed at introducing sparsity during training or removing parameters or kernels from trained networks using a threshold \cite{lebedev2016fast, zhou2016less, wen2016learning, anwar2017structured, polyak2015channel, li2016pruning}. The importance or threshold values are chosen heuristically. Moreover, inducing sparsity through approximate techniques, such as $l_1$-norm constraint, may not be effective. Therefore, the pruning process remains a challenging task due to the manual elimination of kernels or setting up a threshold value for the parameter importance \cite{lebedev2016fast, zhou2016less, wen2016learning, anwar2017structured, polyak2015channel, li2016pruning}.

\begin{figure*} [htpb] 
\begin{center}
\includegraphics[width=1.0\textwidth]{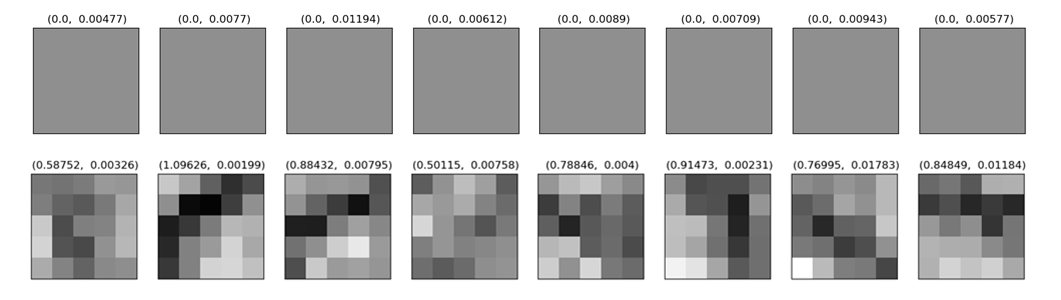}
\caption{Sixteen convolutional kernels (randomly selected out of total one hundred) for the Bayesian eVI CNN trained on the MNIST dataset are presented. On the top of each kernel, Frobenius norm and standard deviation obtained from the kernel's posterior distribution are shown. The top row shows eight kernels with Frobenius norm less than $10^{-7}$, which can be removed without compromising the network's classification accuracy}.  \label{fig:kernels}
\end{center}
\end{figure*}

In this paper, we propose an automated process of network compression by capitalizing on the recent work in Bayesian neural networks \cite{Dera2020radar, dera2019extended}. We leverage uncertainty information available in Bayesian techniques to realize self-compression in convolutional neural networks (CNNs). Consequently, ``less important'' kernels are automatically identified by the Bayesian CNN and can be removed without compromising the performance accuracy. Our proposed Bayesian CNNs perform at the same accuracy level as vanilla CNNs, however, at reduced storage and computational requirements. 

\section{Bayesian Self-Compression} \label{sec:Bayes}

\subsection{Extended Variational Inference}

In the Bayesian framework, the unknown parameters $\Omega$ are treated as random variables, and a prior distribution $\mathcal{}{p(\mathrm{\Omega})}$ is defined over these parameters. Using Bayes' Theorem and available dataset $\cal{D}$, the posterior distribution $p(\mathrm{\Omega}|{\cal{D}})$ and subsequently the predictive distribution are found. Exact Bayesian inference on network parameters is mathematically intractable due to the functional form of these networks and large parameter space \cite{graves2011practical, blundell2015, shridhar2018}.

Variational Inference (VI) is scalable density approximation technique based on optimization which avoids intractable integration operations \cite{graves2011practical, shridhar2018}. VI proposes an approximating distribution $\mathcal{}{q_{\vc{\theta}}(\mathrm{\Omega})}$ and minimize the Kullback-Leibler (KL) divergence between the proposed distribution and the true posterior $p(\mathrm{\Omega}|{\cal{D}})$: 
\begin{equation}\label{eq:1}
\mathbf{KL} (q_{\vc{\theta}}(\mathrm{\Omega})\big|\big|p(\Omega\big|\mathcal{D}))=
\int q_{\vc{\theta}}(\mathrm{\Omega}) \log\frac{  q_{\vc{\theta}}(\mathrm{\Omega})}{p(\mathrm{\Omega}) p( \cal{D} |\mathrm{\Omega})} d \mathrm{\Omega}.
\end{equation}
Using (\ref{eq:1}), an optimization problem in the form of well-known Evidence Lower Bound (ELBO) \cite{graves2011practical} can be formulated as:
\begin{equation}\label{eq:2}
\mathbf{L}(\vc{\theta}) = \mathbf{KL} \left(q_{\vc{\theta}}(\mathrm{\Omega})\big|\big|p(\Omega)\right) 
  -\mathbb{E}_{q_{\vc{\theta}}(\mathbf{\Omega})}\left[\log\left(p(\mathcal{D}|\Omega)\right)\right].
\end{equation}

Recently, Dera \emph{et al.} proposed a VI framework, referred to as the extended Variational Inference (eVI), which propagates the first two moments of the variational posterior through layers of a CNN \cite{Dera2020radar, dera2019extended}. Authors defined the tensor normal distribution over kernels as the prior distribution, and used (\ref{eq:2}) to approximate the posterior distribution of kernels $ \mathcal{N} \left( \vc{\mu} , \sigma^2 \vc{I} \right)$. The estimated $\sigma^2$ provided a measure of confidence or uncertainty attached to the learned kernels during and after the training \cite{dera2019extended, Dera2020radar}. Bayesian eVI CNNs showed improved robustness to noise and adversarial attacks \cite{dera2019extended, Dera2020radar}. In this work, we establish that the Bayesian eVI approach leads to self-compress and improves network performance.

\subsection{Self-Compression with Bayesian eVI}
The proposed method can be considered as a ``self-aware'' compression process for neural networks. During training, a set of kernels is identified as redundant by the Bayesian eVI framework. The redundant kernels are removed through an iterative process, and a compressed network is realized. The Frobenius norm is used to rank each kernel's redundancy or equivalently its importance. The Frobenius norm of a kernel $K$ of size $d \times d$ is defined as:
\begin{equation}
    \left\|K\right\|_F = \sqrt{\text{Tr}\left(KK^T\right)},
\end{equation}
where Tr is the trace operator and $^T$ represents transpose. The Frobenius norm helps us understand the contribution of a kernel to the performance of the Bayesian eVI CNN. 

\begin{figure*}[htpb]
\begin{minipage}[b]{1\linewidth}
  \centering
 \hspace{10mm} \centerline{\includegraphics[width=1\textwidth]{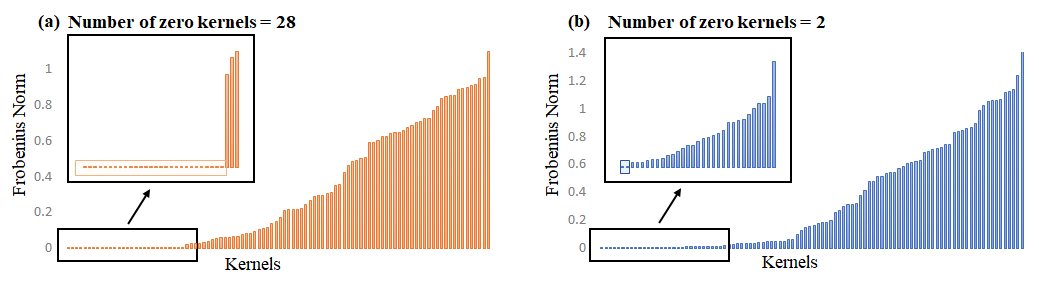}}
 \vspace{-3 mm} 
\end{minipage}
\caption{Frobenius norm of 100 kernels for (a) Bayesian eVI CNN and (b) vanilla CNN trained on the MNIST dataset are presented. The zoomed sub-figures show zero kernels ($\left\| .\right\|_F \le 10^{-7}$). The test accuracy of both CNN was 97\%. We consider that a Bayesian CNN can perform at the same accuracy level with 28 zero kernels, which are a candidate for removal in the next training cycle.}\label{fig:2}
\hspace{5mm} 
\end{figure*}

The MNIST handwritten digits dataset and Fashion-MNIST (F-MNIST) dataset were used to train multiple Bayesian and vanilla CNNs \cite{xiao2017fashion, Deng2012TheMD}. The architecture of all CNNs included one convolution layer followed by the rectified linear unit (ReLU) activation, one max-pooling layer, and one fully-connected layer \cite{Nair:2010}. The kernel size was set to $5 \times 5$ in the convolutional layer of all CNNs. The test accuracy served as the metric to evaluate and compare the performance of Bayesian eVI and vanilla CNNs. In the case of Bayesian CNN, the variances of the predictive distribution provided a measure of the confidence in the classification decision.

We used a different number of kernels in the convolutional layer of both Bayesian and vanilla CNNs. For the MNIST dataset, our first set of CNNs was trained using $N=100$ kernels. For the F-MNIST, our experiments started with $N=128$ kernels. The kernels are then reduced based on the kernel's importance metric computed from the Frobenius norm while the variance information is used to evaluate the network's confidence in the values obtained during training for each kernel.

\section{Results and Discussion}
\subsection{MNIST Dataset}
The first set of experiments included training both Bayesian and vanilla CNNs using $N=100$ kernels. Both CNNs achieved $97\%$ test accuracy. In Fig. \ref{fig:kernels}, we present sixteen randomly selected kernels of a trained Bayesian eVI CNN on MNIST dataset. Fig. \ref{fig:2} shows Frobenius norms of all hundred kernels for both CNNs. We note that for the Bayesian CNN, there are 28 kernels with Frobenius norm equal to zero ($\left\|K\right\|_F \le 10^{-7}$). However, for vanilla CNN, the number is only 2. We refer to the kernels that have $\left\|K\right\|_F \le 10^{-7}$ as \textit{zero kernels}. 

In the case of Bayesian eVI CNN, the average variance $\sigma^2$ was $0.006$ for zero kernels (28 in total) and $0.007$ for non-zero kernels. These low variance values for zero and non-zero kernels indicate the network's high level of confidence in these kernels' values. However, no kernel variance information is available for vanilla CNNs. Through experiments and empirical evidence, we found that the reduction of 28 kernels in the Bayesian eVI CNN and two kernels in the vanilla CNN did not affect the test accuracy.

\begin{figure*}[htpb]
\begin{center}
\hspace{0cm}\includegraphics[width=1\textwidth]{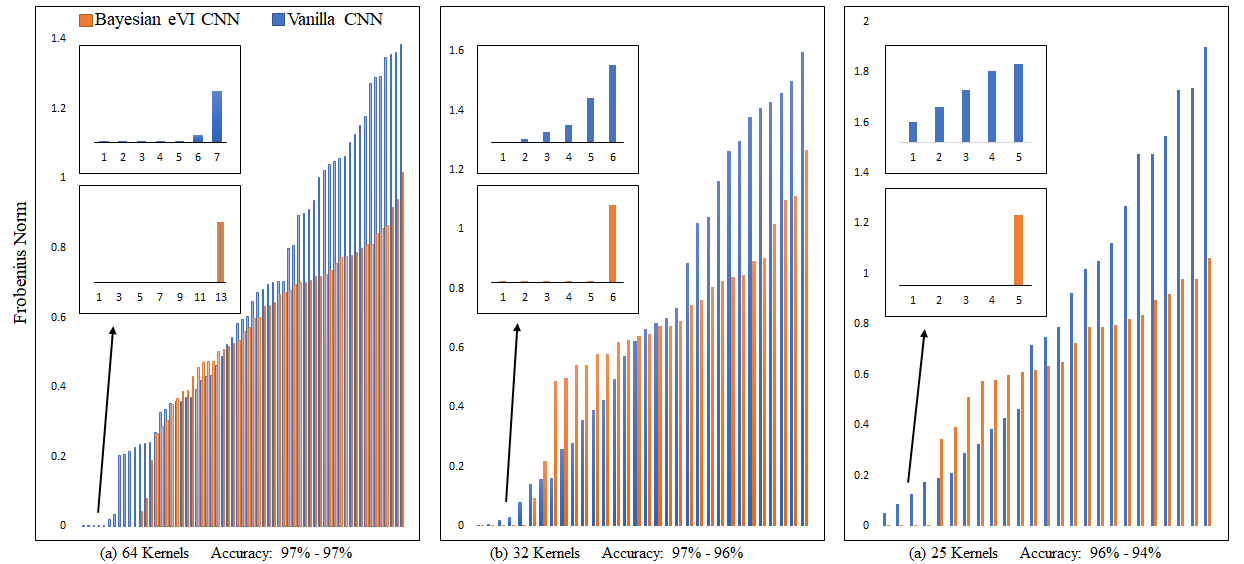}
\vspace{-2mm}
\caption{The Frobenius norms of convolutional kernels for the Bayesian eVI and Vanilla CNNs with $N=64, 32, \text{and}, 25$ for the MNIST dataset are presented. The zoomed sub-figures show kernels with small Frobenius norms. In all cases, Bayesian eVI CNN can discover zero kernels ($\left\| .\right\|_F \le 10^{-7}$). We consider that a Bayesian eVI CNNs provide a principled way to identify and iteratively remove zero kernels, i.e., self-compression.} \label{fig:64_32_25}
\end{center}
\end{figure*}

In the second set of experiments using the MNIST dataset, we trained both the Bayesian eVI and vanilla CNNs with a different number of kernels, i.e., $N=64, 32, \text{and}, 25$. Frobenius norms of kernels for all three cases and both CNNs are presented in Fig. \ref{fig:64_32_25}. In all three cases, Bayesian CNNs have more zero kernels as compared to vanilla CNNs with comparable or higher accuracy. For the case of $N=64$, both CNNs have 97\% test accuracy. For the Bayesian CNN, 11 zero kernels were identified, however, for the vanilla CNN, the number was only 5. Similarly, for the case of N=25, Bayesian CNN has 96\% test accuracy, and vanilla CNN has 94\% test accuracy. We identified 4 zero kernels for the Bayesian CNN and none for the vanilla CNN.     

\subsection{Fashion MNIST Dataset}
\begin{figure*}[htpb]
\begin{center}
\hspace{0cm}\includegraphics[width=1\textwidth]{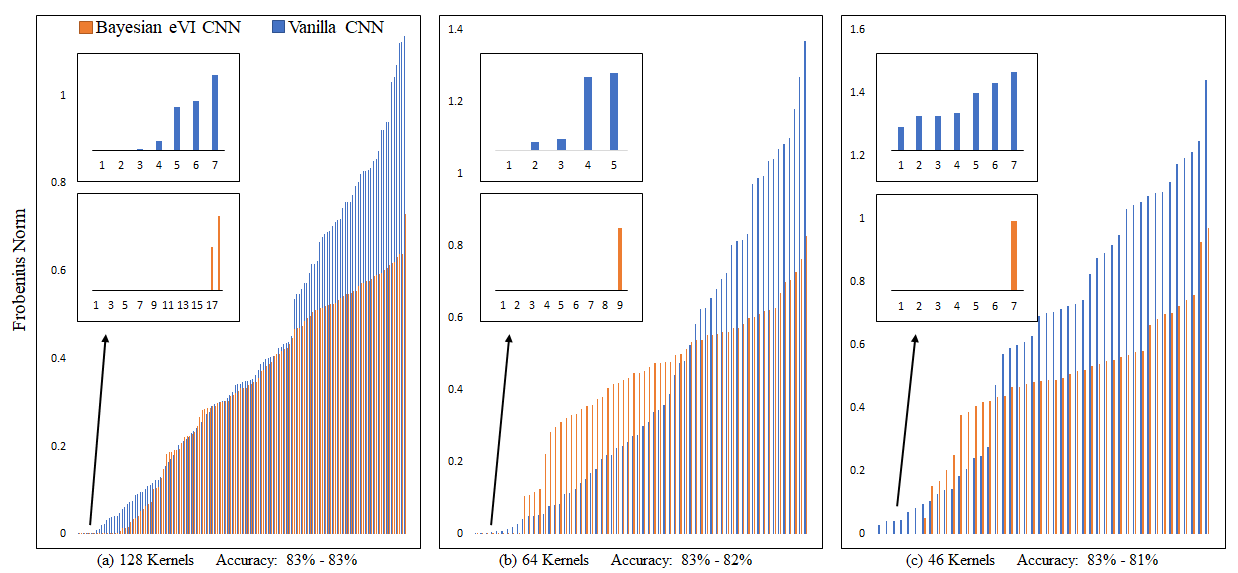}
\vspace{-2mm}
\caption{The Frobenius norms of convolutional kernels for Bayesian eVI and Vanilla CNNs with $N=128, 64, \text{and}, 46$ for the F-MNIST dataset are presented. The zoomed sub-figures show kernels with small Frobenius norm values. We note that for all cases, Bayesian eVI CNN can discover zero kernels and leads to self-compression.} \label{fig:128_64_46}
\end{center}
\end{figure*}

In Fig. \ref{fig:128_64_46}, the Frobenius norms of kernels for both CNNs with $N=128, 64, \text{and}, 46$ are presented. For all three cases, Bayesian eVI CNN achieved the same or better test accuracy; however, only Bayesian eVI CNN was able to identify a set of kernels that could be potentially removed without adversely affecting the test accuracy.  

\subsection{Iterative Compression of CNNs}
\begin{figure}

\begin{center}
\hspace{0cm}\includegraphics[width=0.48\textwidth]{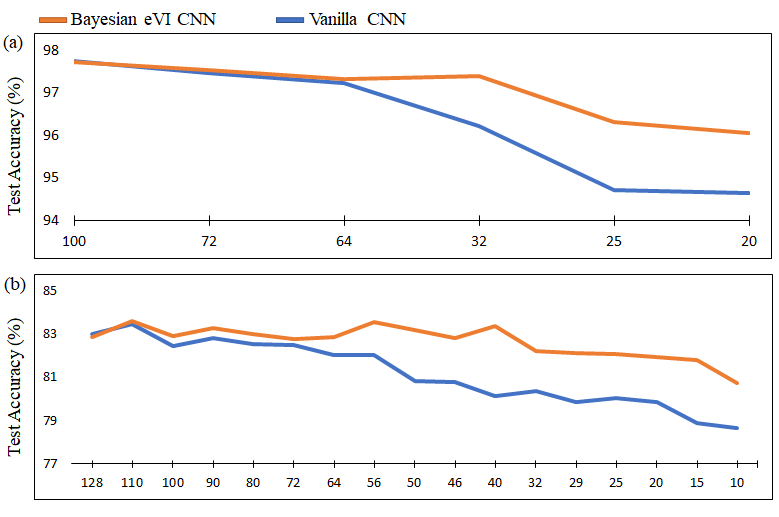}
\vspace{-2mm}
\caption{Test accuracy vs. the number of convolutional kernels $N$ for the Bayesian eVI CNN (orange) and a vanilla CNN (blue) are presented. We note that the Bayesian eVI CNN is able to perform at the same accuracy level as that of vanilla CNN using almost half the number of convolutional kernels (32 vs. 64 for the MNIST dataset and 40 vs. 72 for the Fashion-MNIST dataset). (a) Test accuracy for the MNIST Dataset. (b) Test accuracy for the F-MNIST dataset.} \label{fig:Iterative}
\end{center}
\end{figure}

Based on experiments with MNIST and F-MNIST datasets and Bayesian eVI and vanilla CNNs, we developed an iterative self-compression technique. We used the number of zero kernels identified by the Bayesian eVI CNN as our guide to remove these kernel in an iterative process. The number of identified zero kernels were removed in the next training cycle.  Fig. \ref{fig:Iterative} shows the test accuracy of both Bayesian eVI and vanilla CNNs for a range of the number of kernels $N$ for both datasets. We note that Bayesian eVI CNN retains high accuracy when zero kernels are removed; however, the accuracy of vanilla CNNs starts to drops at a higher rate.

Bayesian eVI CNNs are able to identity zero kernels (i.e., kernels with $\left\| .\right\|_F \le 10^{-7}$) and provide a principled mechanism for the self-compression, i.e., iteratively removing convolutional kernels that are irrelevant. It is important to highlight that the identified zero kernels had low variance values, i.e., the network showed a high level of confidence in these kernels. The capability to identify zero kernels (with high confidence) is exhibited by Bayesian eVI CNNs and is referred to as the \textit{self-compression}.

The self-compression ability of the Bayesian eVI CNNs stems from propagating variational posterior distribution across layers of a neural network during the training process. The availability of the variance information through the variational posterior during the training process allows Bayesian eVI networks to discover important and unimportant (zero) kernels. Vanilla CNNs, on the other hand, owing to lack of information about uncertainty (or variance) tend to use all available kernels to achieve a high accuracy. 

Fig. \ref{fig:Iterative} shows that at the same level of test accuracy, Bayesian eVI CNNs require significantly less number of kernels as compared to vanilla CNNs. The Bayesian eVI CNN performed at the same accuracy level as a vanilla CNN using only half kernels (32 vs. 64 kernels) for the MNIST dataset. Similarly, for the F-MNIST dataset, both Bayesian eVI and vanilla CNNs achieved 83\% accuracy. However, the former used only 40 kernels as compared to 72 for the vanilla CNN. Moreover, the process of discovery of unimportant (zero) kernels is inherently part of Bayesian eVI CNNs. On the other hand, the process of compression or pruning for vanilla CNNs requires manual selection of threshold and possibly a compromise on the test accuracy and performance.  
\vspace{-5mm}
\subsection{The Size of the Parameter Space}
Previously proposed Bayesian approaches for neural networks resulted in a larger parameter space (double the number of parameters as compared to vanilla network) due to the probability distribution defined over the parameters \cite{blundell2015, Dera2020radar}. However, the recently proposed Bayesian eVI CNNs require only one additional variance parameter $\sigma^{2}$ for each convolutional kernel owing to the use of diagonal tensor normal distribution \cite{dera2019extended, Dera2020radar}. A Bayesian eVI CNN has $(d^{2} + 1) N$ parameters as compared to a vanilla CNN that has $d^{2}N$ parameters, where $d$ is the size of a kernel and $N$ represents the total number of convolutional kernels. 

Storage-wise, the additional number of parameters does not appear to be an issue given the \textit{self-compression} capability. For example, in our implementation (with kernel size $d = 5$), the storage for the convolution parameters after compression for both models is shown in table \ref{table:storage}.
\begin{table}[h!]
\caption{Storage (KB) after Compression}\label{table:storage}
\begin{center}
\begin{tabular}{c|c|c}
\hline
  Dataset & Bayesian eVI CNN & Vanilla CNN\\
 \hline
\hline
MNIST & 3.25 & 6.25 \\
\hline
F-MNIST & 4.06 & 7.03\\
\hline
\end{tabular}
\end{center}
\vspace{-5mm}
\end{table}

\section{Conclusion}
\label{sec:conclusion}

The high performance of current machine learning algorithms has associated high computational costs and storage requirements. We exploit the uncertainty information inherent in Bayesian neural networks and propose a method for self-compression of convolutional neural networks. We used the recently proposed extended Variational Inference (eVI) framework that offers advantages of Bayesian neural networks, albeit at a minimal increase in the number of parameters. We showed that Bayesian eVI CNNs were able to identify redundant kernel during training, which can be removed in an iterative process to realize self-compression. The Bayesian eVI CNNs were able to perform at the same accuracy level as that of vanilla CNNs using almost half the number of convolutional kernels. The proposed process of self-compression is part of the training, and resulting Bayesian eVI CNNs are ready for deployment on the edge devices.

\section{Acknowledgement}
\vspace{-2mm}
{\small This work was supported by the National Science Foundation Awards NSF ECCS-1903466. NSF CCF-1527822, NSF OAC-2008690. Giuseppina Carannante is supported by the US Department of Education through a Graduate Assistance in Areas of National Need (GAANN) program Award Number P200A180055. We are also grateful to UK EPSRC support through EP/T013265/1 project NSF-EPSRC: ShiRAS. Towards Safe and Reliable Autonomy in Sensor Driven Systems.

\bibliographystyle{IEEEbib}
\bibliography{strings,ref_prun}}

\end{document}